\documentclass{article}

\usepackage{arxiv}
\usepackage{amssymb}
\usepackage{amsmath}
\usepackage{graphicx}
\usepackage{multirow}
\usepackage{setspace}
\usepackage{color}

\usepackage[utf8]{inputenc} 
\usepackage[T1]{fontenc}    
\usepackage{hyperref}       
\usepackage{url}            
\usepackage{booktabs}       
\usepackage{amsfonts}       
\usepackage{nicefrac}       
\usepackage{microtype}      
\usepackage{lipsum}		
\usepackage{graphicx}
\usepackage{natbib}
\usepackage{doi}

\title{\emph{RF-LighGBM}: A probabilistic ensemble way to predict customer repurchase behaviour in community e-commerce}

\author{ \href{https://orcid.org/0000-0002-8370-556}{\includegraphics[scale=0.06]{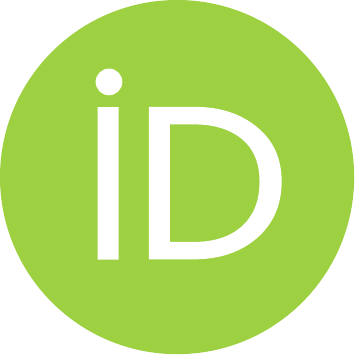}\hspace{1mm}Liping Yang} \\
	School of Economics and Management\\
	Beijing University of Chemical Technology\\
	Beijing, China. \\
	\texttt{lipingphd@163.com} \\
	\And
	{Xiaxia Niu, Jun Wu} \\
	School of Economics and Management\\
	Beijing University of Chemical Technology\\
	Beijing, China. \\
}

\renewcommand{\shorttitle}{\textit{arXiv} Template}

\hypersetup{
pdftitle={A template for the arxiv style},
pdfsubject={q-bio.NC, q-bio.QM},
pdfauthor={David S.~Hippocampus, Elias D.~Striatum},
pdfkeywords={First keyword, Second keyword, More},
}

\begin{document}
\maketitle
\setstretch{1.23} 
\begin{abstract}
	It is reported that the number of online payment users in China has reached 854 million; with the emergence of community e-commerce platforms, the trend of integration of e-commerce and social applications is increasingly intense. Community e-commerce is not a mature and sound comprehensive e-commerce with fewer categories and low brand value. To effectively retain community users and fully explore customer value has become an important challenge for community e-commerce operators. Given the above problems, this paper uses the data-driven method to study the prediction of community e-commerce customers' repurchase behaviour. The main research contents include 1. Given the complex problem of feature engineering, the classic model RFM in the field of customer relationship management is improved, and an improved model is proposed to describe the characteristics of customer buying behaviour, which includes five indicators: recent consumption time ($R$), purchase frequency ($F$), total purchase amount ($M$), customer relationship duration ($S$) and average transaction time interval ($T$). 2. In view of the imbalance of machine learning training samples in SMOTE-ENN, a training sample balance using SMOTE-ENN is proposed. The experimental results show that the machine learning model can be trained more effectively on balanced samples. 3. Aiming at the complexity of the parameter adjustment process, an automatic hyperparameter optimization method based on the TPE method was proposed. Compared with other methods, the model's prediction performance is improved, and the training time is reduced by more than 450\%. 4. Aiming at the weak prediction ability of a single model, the soft voting based RF-LightgBM model was proposed. The experimental results show that the RF-LighTGBM model proposed in this paper can effectively predict customer repurchase behaviour, and the $F_1$ value is 0.859, which is better than the single model and previous research results.
\end{abstract}


\section{Introduction}
With the rapid development of e-commerce, users can use real-time online services through mobile terminals to consume offline stores. Information provided by stores can also be used to transfer offline consumption to online consumption through mobile terminals. The virtual platform and physical retail combination forms online to offline (O2O) or offline to online shopping activities and consumption behaviours, such as Dianping, Meituan and other group buying platforms. According to the 47th Statistical Report on The State of Internet Development in China (2021), by December 2020, the scale of Online payment users in China has reached 854 million, and the trend of integration of e-commerce and social applications is increasingly vital. For e-commerce enterprises, customers' repurchase behaviour (retention) is an essential source of profit. At the same time, most new customers are one-off traders \citep{liu2016repeat}. The mature and sound comprehensive e-commerce of community e-commerce has the disadvantages of fewer categories and low brand value. To effectively retain community users and fully explore customer value has become an important challenge for community e-commerce operators.

Existing researches mainly predict customer purchase behaviour from contract scenario and non-contract scenario. Contract scenario mainly refers to the continuous purchase of products or services by customers through establishing contractual relations with enterprises, such as continuous subscription of newspapers and periodicals. Non-contract scenario refers to the independent purchase behaviour that the customer and the enterprise do not sign a purchase contract. Most of the community e-commerce customers' purchases belong to non-contract scenarios. The prediction of customer purchase behaviour in the non-contract scenario is the core research content of Customer Relationship Management (CRM)(\citep{schmittlein1987counting},\citep{fader2005rfm},\citep{abe2009counting}). The repurchase behaviour prediction of community e-commerce customers studied in this paper refers to the behaviour prediction of customers who have purchased a specific product on a community e-commerce platform in a certain period to purchase the product again in the future.

At present, the research methods of customer repeat purchase behaviour in non-contract scenarios can be divided into two categories: model-based method and data-driven method. The model-based approach was the first to be applied. The Pareto/NBD model proposed by \citep{schmittlein1987counting} is the most classical probabilistic model to depict customer purchase behaviour in non-contractual scenarios, and a series of works have been carried out based on it. \citep{fader2005counting} proposed a new model BG(beta-geometric)/NBD model based on the Pareto/NBD model from the perspective of simplifying model assumptions. This model considers that the customer does not bring value to the enterprise from the last purchase to the loss. It changes customer loss from Pareto distribution to Geometric distribution and determines whether it occurs immediately after each purchase. The results show that the BG/NBD model is much better than the Pareto/NBD model in computation complexity and time. \citep{abe2009counting} believes that the assumption that customer churn rate and repurchase rate are independent is difficult to establish in actual situations, so the assumption that customer churn rate and repurchase rate are independent of each other is changed into a correlation. However, the above methods rely on rigorous mathematical deduction and probability distribution when studying customer purchase behaviour prediction, which is not easy in practical application.

With the explosion of enterprise data and the rise of machine learning methods, data-driven methods are paid attention to. The data-driven method does not need to make an accurate analysis of probability distribution. By mining and analyzing customer purchase records recorded in enterprise information systems, knowledge hidden behind behaviours can be obtained, which is more suitable for the characteristics of the current big data era. \citep{liu2016repeat} used a variety of machine learning models to analyze the new customer data acquired on Tmall's "Double 11", constructed 1364 features and obtained relatively good results. \citep{shen2020reconciling}, based on the Alibaba e-commerce platform, uses the tree model with interpretation methods  from Factor Importance, Partial dependence plots and Decision Rules to predicts repeat buyers. However, the above literature has shortcomings in predicting customer purchasing behaviour: 1) Feature engineering (the modelling of consumer purchasing behaviou) is very complex. 2) Most studies are based on e-commerce platforms with rich categories, such as Tmall e-commerce. This kind of platform has the characteristics of a complete system, comprehensive record, and a large number of users, which many community e-commerce platforms do not possess. 3) The parameter adjustment process is complicated. Machine learning model hyperparameters have a significant influence on model prediction performance. Grid search and other existing methods cannot distinguish between model hyperparameters and model performance, and the model training process is time-consuming and inefficient.

The primary purpose of this paper is to overcome the shortcomings of the above methods and provide more accurate and intelligent decision support for community e-commerce operators. The main innovation points include four aspects :(1) given the complex problem of feature engineering, starting from the previous model-based methods and referring to the idea of the classical model, customer behaviour is modelled. Furthermore, the input features are tested by various methods to avoid redundant features and irrelevant features. (2) In view of the imbalance of machine learning training samples in SMOTE-ENN, this paper puts forward a method for balancing training samples in predicting customer SMOTE-ENN. The experimental results show that the machine learning model can be trained more effectively on balanced samples. (3) Aiming at the complex parameter adjustment process problem, an automatic hyperparameter optimization method is proposed based on the TPE method. By setting the objective optimization function of each machine learning model, the knowledge between model hyperparameters and model prediction performance can be formed. Compared with other methods, model training efficiency and model prediction performance can be significantly improved. (4) Aiming at the problem of low prediction ability of a single model, a soft voting method based RF-LightGBM model is proposed. Experimental results show that the RF-LightGBM model proposed in this paper can effectively predict customer repurchase behaviour and is superior to the single model and previous research results.

The following sections of this paper are arranged as follows: Section 1 mainly introduces the theoretical basis of the method proposed in this paper; Section 2 introduces the community e-commerce platform and model evaluation indicators applied by the method in this paper. The experimental results are analyzed and discussed in Section 3. The conclusion of this paper is presented in section 4.

\section{Theory and Method}
\label{sec:Theory and Method}
\begin{figure}
    \centering
    \fbox{\includegraphics{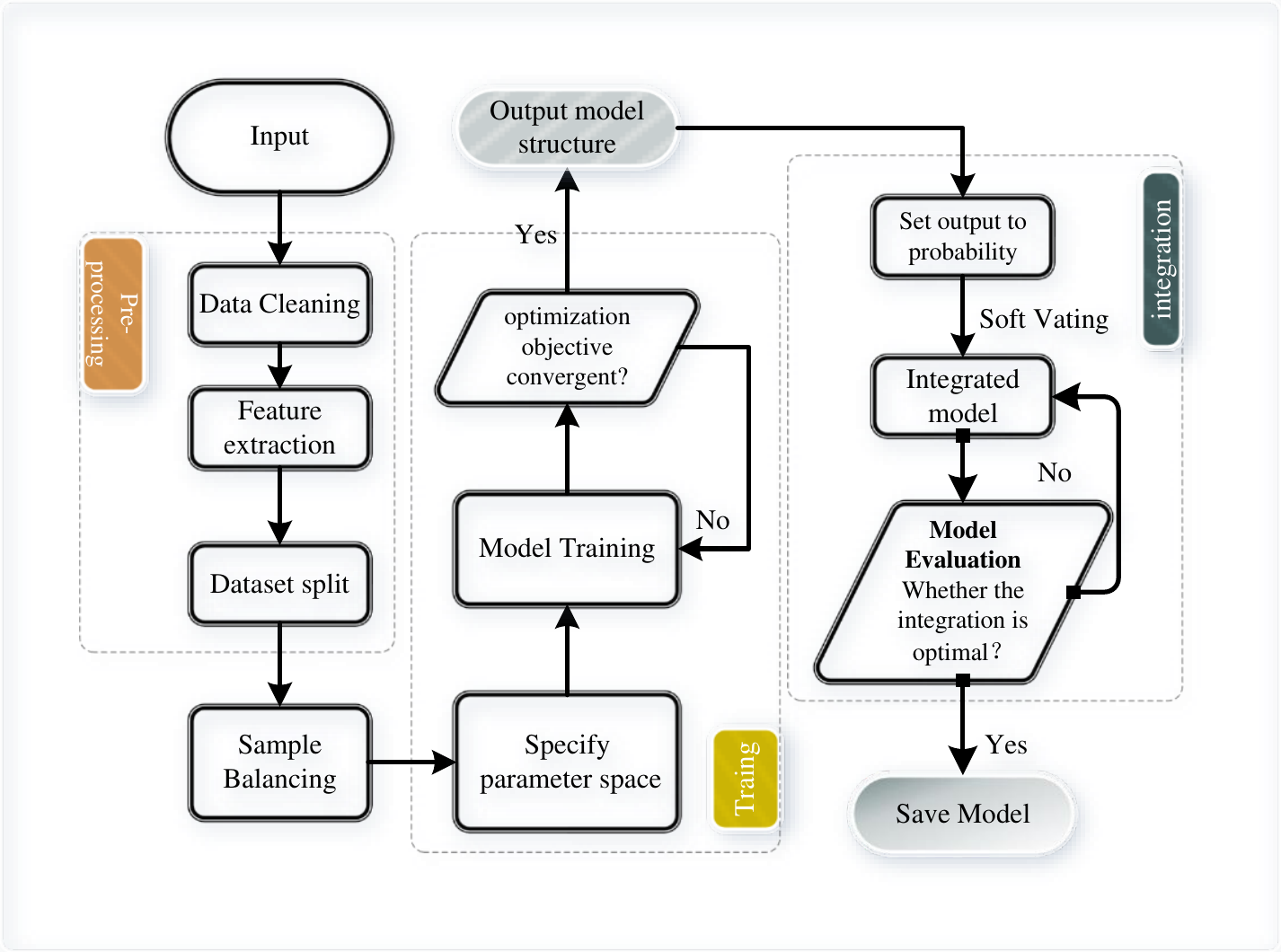}}
    \caption{Overall flow chart of methodology}
    \label{fig:fig.1}
\end{figure}

The research methodology adopted in this paper can be summarized as Fig.\ref{fig:fig.1}
, roughly divided into three parts: data preprocessing, optimization training and model integration. In data preprocessing, the original data is cleaned by eliminating invalid attributes and deleting missing values. Secondly, the corresponding features are extracted to construct the data set. Finally, the data set is divided into the training set, verification set and test set. In the optimization training part, the space range of the hyper-parameters of the machine learning model is specified first, and then the TPE method is used to train the model iteratively on the training set until the objective function converges to output the optimal model structure. Finally, the model integration part first specifies the output of each model as the classification probability and then use the soft vote method in the integration method to integrate multiple model combinations, output the optimal integration model.

\subsection{Modeling Consumer Repurchase Behavior}
RF (recency-frequency) and RFM (recency-frequency-monetary) models are the two main branches in the early modelling of customer repurchase behaviour. The Pareto/NBD model proposed by \citep{schmittlein1987counting} attempts to model the interval between two purchases by customers and the interval between the last purchase by customers and the current observation time point. The subsequent work of \citep{fader2005counting} and \citep{abe2009counting} can also be summarized into the RF model because it only considers two kinds of purchase information: the frequency of purchase and the interval between the last purchase and the end of the observation period. \citep{fader2005rfm} further proposed the RFM model, taking into account the currency information purchased by customers, and presented the customer life cycle value calculation method. \citep{zhang2015predicting} took the concentration degree of customer purchase time into consideration and illustrated the importance of this improvement through group comparison.

Compared with full-category e-commerce platforms, community e-commerce tends to have the following characteristics \citep{wu2021user}: 1) Relatively fixed user groups; 2) Consumer goods are relatively undiversified; 3) Repeat purchase features are apparent. The literature research (\citep{wu2021user},\citep{wu2020empirical}) proves that the RFM model and its improved model can be well applied to customer value identification of community e-commerce platforms. As a classical CRM model, this model can describe customer profiles in multiple dimensions and is also an effective method to predict the repeated purchase behaviour of consumers. Based on the literature \citep{wu2021user}, the following five characteristics are adopted to characterize the purchasing behaviour of users:

\textbf{R:} The interval between the last consumption time of the customer and the end of the observation period.
\begin{equation}
    R = T_{last\_time}-T_{plast\_time}
\end{equation}
Where, $T_{last\_time}$ is the end of the reference period, $T_{plast\_time}$ is the last transaction time of the customer's order for the product within the reference period.

\textbf{F: }Purchase frequency of the product during the observation period.

\textbf{M: }The total amount of the product purchased by the customer.
\begin{equation}
    M=\sum_{i}^{n}{M_i}
\end{equation}
Where, $n$ refer to the total number of customer purchases during the period, $M$ refers to amount of a customer's single purchase.

\textbf{S: }The interval between the first transaction and the last transaction occurred during the customer's observation period.
\begin{equation}
    S=T_{plast\_time}-T_{pfirst\_time}
\end{equation}
Where, $T_{plast\_time}$ represents the transaction time of the last customer order for the product within the observation period, $T_{pfirst\_time}$ refers to the transaction time of the customer's first order for the product within the reference period.

\textbf{T:} represents the average transaction time interval of a customer within a certain period.
\begin{equation}
    T=\frac{S}{F}
\end{equation}
Generally, the smaller the $R$ index value is, the more recent the customer has purchased the product, and the higher the probability of repurchase is. $F$ and $M$ can represent customer loyalty to goods. $S$ can represent the duration of the relationship between the customer and the product. The longer the duration of the relationship, the higher the probability of the customer repurchasing the product. $T$ can represent customers' consumption habits. Generally, the smaller $T$ is, the more likely users are to repurchase in the next period.

\subsection{Sample Balancing}
In general, the proportion of consumers who make repeated purchases in stores or platforms is smaller than that of consumers who make single purchases, which leads to the problem of unbalanced sample categories, which often leads to over-fitting of the model and thus affects the classification effect. For example, in only 5\% of the re-purchase prediction of < customer, product >, if the unknown samples guessed by the model are all negative examples, that is, there is no re-purchase behavior, 95\% accuracy can be achieved, which is easy to cause lazy model training, which is not conducive to the construction of prediction model. At present, data level methods to solve data imbalance classification include over-sampling and under-sampling methods. SMOTE (Synthetic Minority Over-piling Technique) \citep{he2009learning},\cite{batista2004study} takes the advantage of self-help method and the k-nearest neighbor method to generate new data similar to the observation of a few classes based on feature space to reduce the error of the classifier. The specific steps are as follows:

Let $A$ represent minority class, take $X_i\in{A}$ calculate its distance to all samples in the sample set $A$ using Euclidean distance as the standard, get k-nearest neighbor samples of $X_i$, and select $A$ sample randomly from this nearest neighbor sample, that is, $X_ij$ (j=1,2,…,n); In $X_i$ and $X_ij$ (j=1,2,…,n) to construct a new minority sample $Y_j$. This process is equivalent to randomly selecting one sample on the line between two samples as (5).
\begin{equation}
    Y_j=X_i+rand(0,1)\times{X_{ij}-X_i}
\end{equation}

\begin{figure}
    \centering
    \fbox{\includegraphics[width=0.75\textwidth]{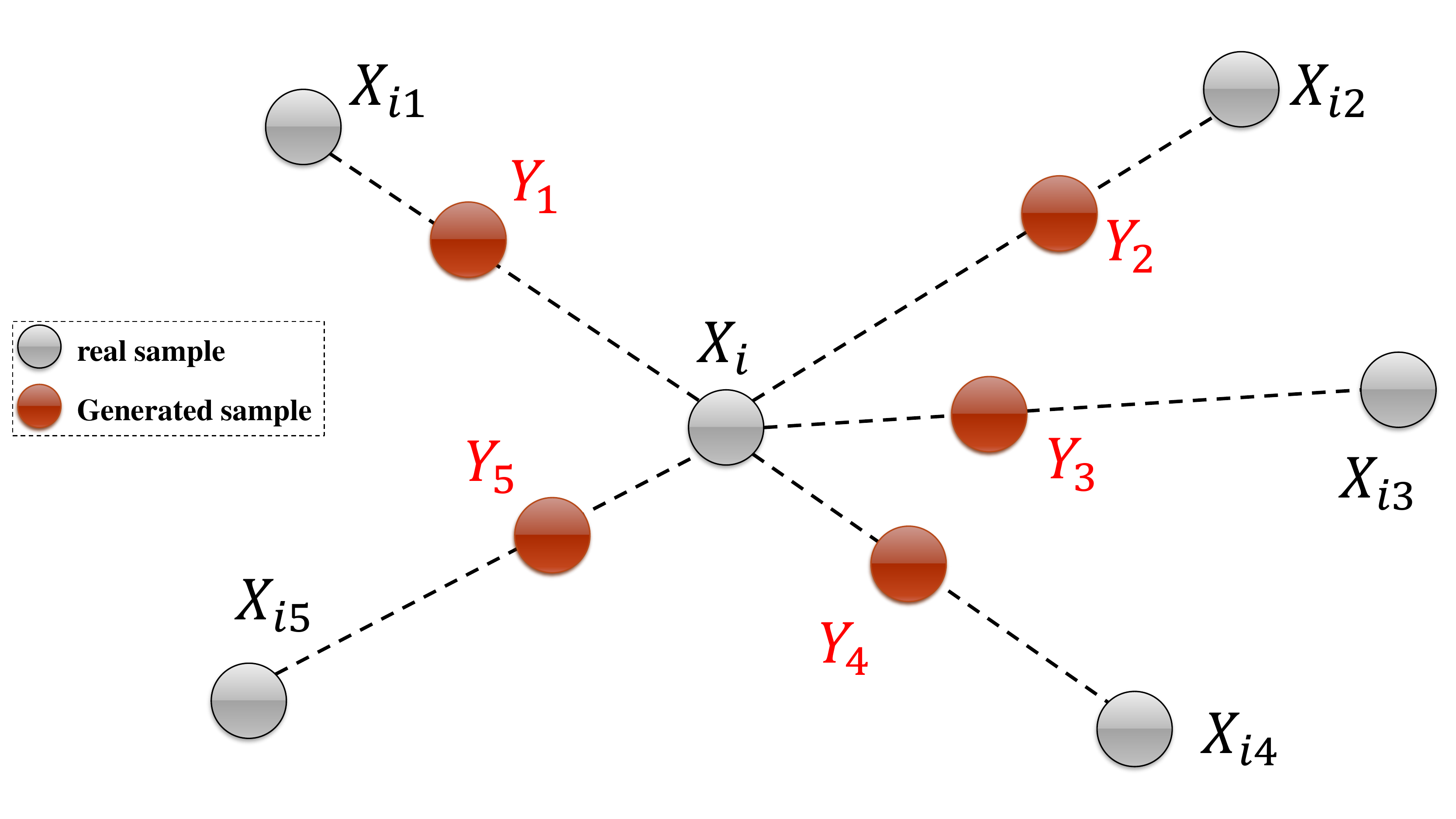}}
    \caption{Overall flow chart of methodology}
    \label{fig:fig.2}
\end{figure}

Where $rand(0,1)$ represents a random number within the interval $(0,1)$. The schematic diagram of the synthesized sample is shown in Fig.\ref{fig:fig.2}.
The SMOTE-ENN method is a combination of SMOTE and ENN methods, which has been proved to be effective in solving the problem of sample imbalance \cite{batista2004study}. The specific steps of using the SMOTE-ENN method for sample balance are as follows: 1) Use SMOTE method for data expansion of small sample class. 2) The KNN method (K=5 here) is used to predict the generated new sample data. If the predicted result is different from the actual category label, the sample will be excluded, and if the same, the sample will be retained.

\subsection{Model Selection}
Integration method refers to a method to solve the inherent defects of a single model or a set of model parameters so as to integrate more models, learn from each other and avoid limitations. Compared with other methods, integrated models often show better predictive performance. Therefore, this paper chooses three mainstream integration models (namely, random forest, XGBoost, LightGBM) as sub-models.

Random Forest is identified as one of the most promising classifiers in many studies \citep{fernandez2014we}. It is the generalization of the decision tree model, which builds the forests based on the ensemble of decision trees called the ‘bagging method’ to get accurate and consistent results. Using the random selection of features, each decision tree is trained on bagged data. The insight into the decision process can be understood by feature importance based on the performance of the model. The algorithm steps can be summarized as follows \citep{breiman2001random}:
Given training vector $x_i$ and label vector $y_i$, use $Q$ to represent the data on m node. Each candidate group $\theta=(j,t_m)$ contains a feature $j$ and a threshold $t_m$. Divide data into $Q_{left}(\theta)$ and $Q_{right}(\theta)$(if $x\leq{t_m}$, $Q$ belongs to $Q_{left}(\theta)$, else $Q_{right}(\theta)$). Use the impurity function $H$ to calculate the impurity at $m$:
\begin{equation}
    G(Q,\theta)=\frac{n_{left}}{N_m}H(Q_{left}(\theta))+\frac{n_{right}}{N_m}H(Q_{right}(\theta))
\end{equation}

Select parameters that minimize impurity $\theta^{8}=argmin_{\theta}G(Q,\theta)$, recurse over $Q_{left}(\theta^{*})$ and $Q_{right}(\theta^{*})$ until you reach the maximum allowable depth.

The specific process of the random forest method can be expressed as follows: 1) Generate M training sets by bootstrap method; 2) For each training set, a decision tree is constructed. When nodes are searching for features and splitting, instead of finding the one that can maximize the index (such as information gain) for all features, some features are randomly selected, and the optimal solution is found among the selected features and applied to nodes for splitting.

XGBoost algorithm \citep{chen2016xgboost} uses an approximate method to solve the problem that traditional Gradient Tree Boosting objective function is difficult to optimize in Euclidean space. First, rewrite the original objective function $obj$:
\begin{equation}
    Obj^{(s)}=\sum_{i=1}^{n}{L(y_i,\hat{y}_i^{(s-1)}+f_x(x_i))}+\Omega(f_s)
\end{equation}
Where $\hat{y}_i^{(s-1)}$ is predicted value of the $s-1$ round sample $x_i$, $f_s(x_i)$ is a new sub-model of the s round training. The objective function is approximated and simplified by introducing Taylor's formula. Taylor's formula is a formula that uses the information of a certain point of a function to describe its nearby value. If the function curve is smooth enough, a polynomial can be constructed through the derivative value of a certain point to approximate the value of the function in the field of this point. Here, we only take the two orders of Taylor's demonstration, which are defined as follows:
\begin{equation}
    f(x+\vartriangle{x})\cong{f(x)+f'(x)\vartriangle{x}+\frac{1}{2}f''x\vartriangle{x}^2}
\end{equation}

Take $\hat{y}_i^{(s-1)}$ in the above equation as $x$ and consider $f_s(x_{i})$ as $\vartriangle{x}$. Then, Taylor expansion can be performed on the objective function:
\begin{equation}
    Obj^{(s)}\cong{\sum_{i=1}^{n}{[L(y_i,\hat{y}^{(s-1)})+g_if_s(x_i)+\frac{1}{2}h_if_s^2(x_i)]}+\Omega{f_s}}
\end{equation}
Where, $g_i$ is the one-step statistics of the loss function, $h_i$ is the two-step statistics of the loss function, as follows:
\begin{equation}
    g_i=\frac{\partial{L(y_i,\hat{y}^{(s-1)})}}{\partial{\hat{y}^{(s-1)}}}
\end{equation}
\begin{equation}
    h_i=\frac{\partial^2{L(y_i,\hat{y}^{(s-1)})}}{\partial{\hat{y}^{(s-1)}}}
\end{equation}

Because the constant term doesn't affect the optimization result, we can further simplify $Obj^{s}$, remove the constant term $L(y_i,\hat{y}^{(s-1)})$, and take $\Omega(f_s)$ in it, we obtain:
\begin{equation}
    Obj^{s}=\sum_{i=1}^{n}{[g_if_s(x_i)+\frac{1}{2}h_if^2_{s}(x_i)]+\gamma{T}+\frac{1}{2}\lambda\sum_{j=1}^{T}{\omega_j^2}}
\end{equation}

Therefore, XGBoost model greatly improves the training speed and prediction accuracy.

The LightGBM was originally developed by researchers from Microsoft and Peking University \citep{ke2017lightgbm} to address the efficiency and scalability issues with GBDT and XGBoost when applied to problems with high-dimensional input features and large data sizes. It is reported that LightGBM outperforms other gradient boosting methods such as GBDT and XGBoost in terms of training speed and predictive accuracy since LightGBM incorporates two innovative techniques: Gradient-based One-Side Sampling (GOSS) and Exclusive Feature Bundling (EFB).

Suppose a dataset has $n$ instances $S=\{(x_1,y_1),(x_2,y_2),\ldots,(x_n,y_n)\}$, where $\{x_1,x_2,\ldots,x_n\}$ and $\{y_1,y_2,\ldots,y_n\}$ are independent and dependent variables, respectively. The predicted value of GBDT (($f_x$)) is the sum of the outputs of a set of decision tree models ($h_t(x)$):
\begin{equation}
    f(x)=\sum_{t-1}^{T}{h_t(x)}
\end{equation}
Where $T$ is the number of decision trees. Fitting a GBDT model is to find an approximation function $\hat{f}$ which minimizes the loss function $L(y,f(x))$ as shown in Eq. (14):
\begin{equation}
    \hat{f}=\arg\min_{f}{E_{y,s}L(y,f(x))}
\end{equation}

Instead of splitting the internal nodes of each tree using information gain like traditional GBDT, LightGBM splits the internal nodes using the Gradient-based One-Side Sampling (GOSS) method. Specifically, samples with larger absolute values of gradients (i.e., top $\alpha$*100\%) are selected as subset $A$ while the remaining samples with smaller gradients are randomly chosen to form subset $B$(i.e.,$b$*(1-$\alpha$)*100\%). Hence, the samples are split according to the variance gain $V_j(d)$on $A\cup{B}$:
\begin{equation}
    V_j{(d)}=\frac{1}{n}{[\frac{(\sum_{x_i\in{A_l}}{g_i+\frac{1-a}{b}\sum_{x_i\in{B_l}}{g_i}})^2}{n_l^j(d)}+\frac{(\sum_{x_i\in{A_r}}{g_i+\frac{1-a}{b}\sum_{x_i\in{B_r}}{g_i}})^2}{n_r^j(d)}]}
\end{equation}
Where, $A_l={x_i\in{A}:x_{ij}\leq{d}},A_r = {x_i\in{A}:x_{ij}>{d}}, B_l={x_i\in{B}:x_{ij}\leq{d}}, B_l={x_i\in{B}:x_{ij}> {d}}$ and $g_i$ denotes the negative gradients of the loss function for the LightGBM outputs in each iteration.

Besides employing GOSS for sampling, LightGBM also uses the Exclusive Feature Bundling (EFB) to speed up the training process without losing accuracy. Many applications have high-dimensional and sparse input features that are also mutually exclusive (i.e., these features cannot be nonzero at the same time). EFB can bundle these features together into a single feature bundle. The feature histograms from these feature bundles as well as those from individual features can be constructed using a feature scanning algorithm.

In summary, LightGBM is an emerging ensemble machine learning (EML) method that uses GOSS to split internal nodes based on variance gain and EFB for input features dimension reduction. As a decision tree based model, LightGBM has an additional advantage of being robust against multicollinearity (Kotsiantis, 2013) [14]. Hence, including correlated independent variables, which is quite common in safety data, in LightGBM model is not an issue.

\subsection{Hyperparameter Optimization}
Model optimization is one of the important links in machine learning. Most branches of machine learning theory are devoted to model optimization \citep{lujan2018design}. Compared with other methods, automatic hyper-parameter tuning can form the knowledge between the parameters and the model, to reduce the number of tests and improve the efficiency of the algorithm when searching for the optimal hyper-parameter. 

TPE algorithm \citep{bergstra2011algorithms} is an optimization method based on a sequential model. This method converts the hyper-parameter space nonparametric density distribution, the modeling process of $p(x|y)$. There are three conversion modes: uniform distribution to truncated Gaussian mixture distribution, logarithmic uniform distribution to exponential Gaussian mixture distribution, and discrete distribution to reweighted discrete distribution. By using different observed values in the nonparametric density ($x^1,x^2,…,x^k$). TPE can use learning algorithms of different densities. Its density is defined as Eq.(16).
\begin{equation}
    p(x|y)=\left\{
    \begin{aligned}
    l(x) & , & if \ y<{y^*} \\
    g(x) & , & if \ y\geq{y^*}
    \end{aligned}
    \right.
\end{equation}
$l(x)$ consists of a density where the objective function $F(x)$ of the observed value \{$x^i$\} is less than $y^*$, while $g(x)$ consists of the density of the objective function F(x) of the observed value \{$x^i$\} greater than or equal to $y^*$. TPE uses $y^*$ as the sub-site $\gamma$ of the observed value $y$. Through maintenance ordered list of observation data in the observational domain H, each iteration of TPE algorithm running time can be in $|H|$ and the characteristics of the optimized dimensions of the linear scaling. In this case, expect improvement ($EI$) is:
\begin{equation}
    EI_{y^*}(x)=\int_{-\infty}^{\infty}{y^*-y}p(y|x)dy=\int_{-\infty}^{y^*}(y^*-y)\frac{p(x|y)p(y)}{p(x)}dy
\end{equation}

Finally, by constructing $\gamma=p(y<y^*)$ and $p(x)=\int{p(x|y)p(y)dy=\gamma{l(x)+(1-\gamma)g(x)}}$, (18) is easy to obtain.
\begin{equation}
    EI_{y^*}(x)=(\gamma+\frac{g(x)}{l(x)}(1-\gamma))^{-1}
\end{equation}
Thus, each iteration will return an $x^*$ with the maximum $EI$.

\subsection{Model Integration}
Algorithm integration is to form a new combination model by integrating the learning results of several single algorithms so as to improve the final learning accuracy of the algorithm. At present, there are many algorithms fusion methods, such as the Bagging method used by random forest algorithm, Boosting algorithm used by LightGBM algorithm, etc.

Integrated algorithms usually achieve better generalization performance than single algorithms, which can be explained from the following three intuitive perspectives:

1) Data perspective: The information provided by a sample set is not enough to enable the learning algorithm to pick out the correct hypothesis, but the numerous hypotheses selected by the algorithm reach a certain accuracy. By integrating the optimal hypotheses, the learning algorithm can approach the only correct hypothesis in the space.

2) Algorithm perspective: The correct hypothesis describing a sample set may not be in the hypothesis space of an algorithm. By integrating multiple hypotheses in this hypothesis space, this hypothesis space can be expanded to approach the correct hypothesis.

3) Calculation Angle: Many algorithms only perform a local search in the hypothesis space and often miss the most hypothesis and fall into local extremums. The first-year algorithm in the fusion algorithm searches from different starting points and can approximate the optimal hypothesis better.

The random forest algorithm, XGBoost algorithm and LightGBM algorithm in this paper are all integrated algorithms with different strategies. Further integration of them is expected to achieve better prediction performance. The soft voting method determines the category of test samples by calculating the average probability of each model for each category and comparing the sizes. In the following article, the soft voting method is used to integrate the model. The detailed steps can be expressed as follows:

1.Specify that the output values of all three models are probabilistic, as Eq.(19).
\begin{equation}
    Model(P(x|y))=\sum_{i=1}^{N_{tree}}{h_i(P(x|y))}
\end{equation}
Where, $N_tree$ is the total number of decision trees, $h_i$ is $i$ decision tree, Model represents model selected in this paper (Random Forests, XGBosot, LightGBM), P(x|y) represent probability of predicting sample $x$ belongs to class $y$.

2.Assign the same weight to the submodel, use Soft Voting method for integration based on its predicted probability, and its mathematical expression is as follows:
\begin{equation}
    P_{SoftVoting}=(P_{model1}+P_{model2})/2
\end{equation}
\begin{equation}
    Result=\left\{
    \begin{aligned}
    1 & , & if \ P_{SoftVoting}\geq{threshold} \\
    0 & , & if \ P_{SoftVoting}<{threshold}
    \end{aligned}
    \right.
\end{equation}
Where, $P_{SoftVoting}$ refers to the prediction probability of the soft vote integration model, $P_{modeli}$ represents submodel, $Result$ is the prediction result, and 1 represents the users who have repurchased, 0 represents the users who have not. $threshold$ is classification threshold.

\section{Experiment}
\subsection{Data Description}
\begin{figure}
  \begin{minipage}[t]{0.5\linewidth}
    \centering
    \fbox{\includegraphics[scale=0.4]{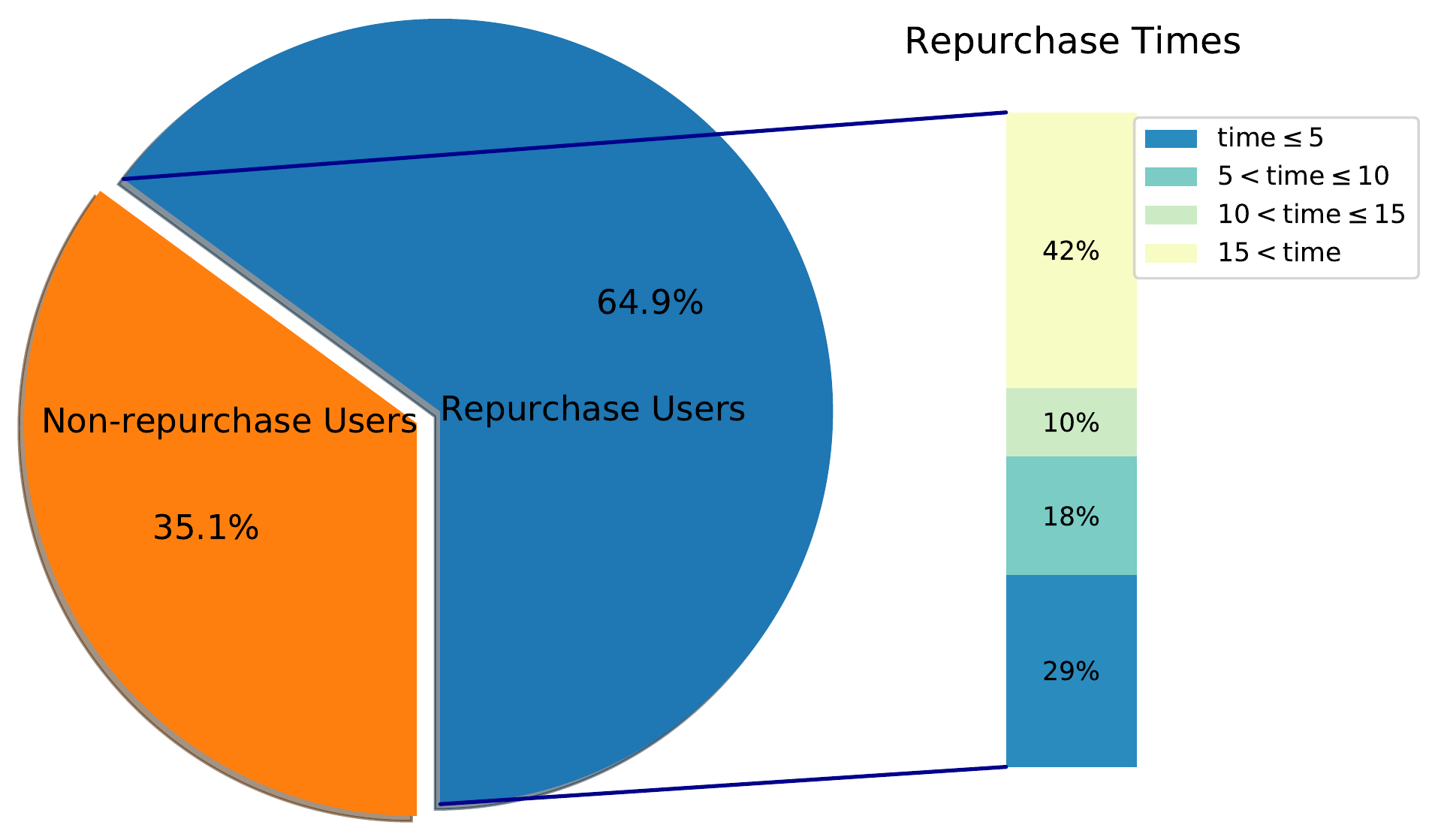}}
    \caption{Customer Purchase Behavior Description}
    \label{fig:fig.3}
  \end{minipage}%
  \begin{minipage}[t]{0.5\linewidth}
    \centering
    \fbox{\includegraphics[scale=0.4]{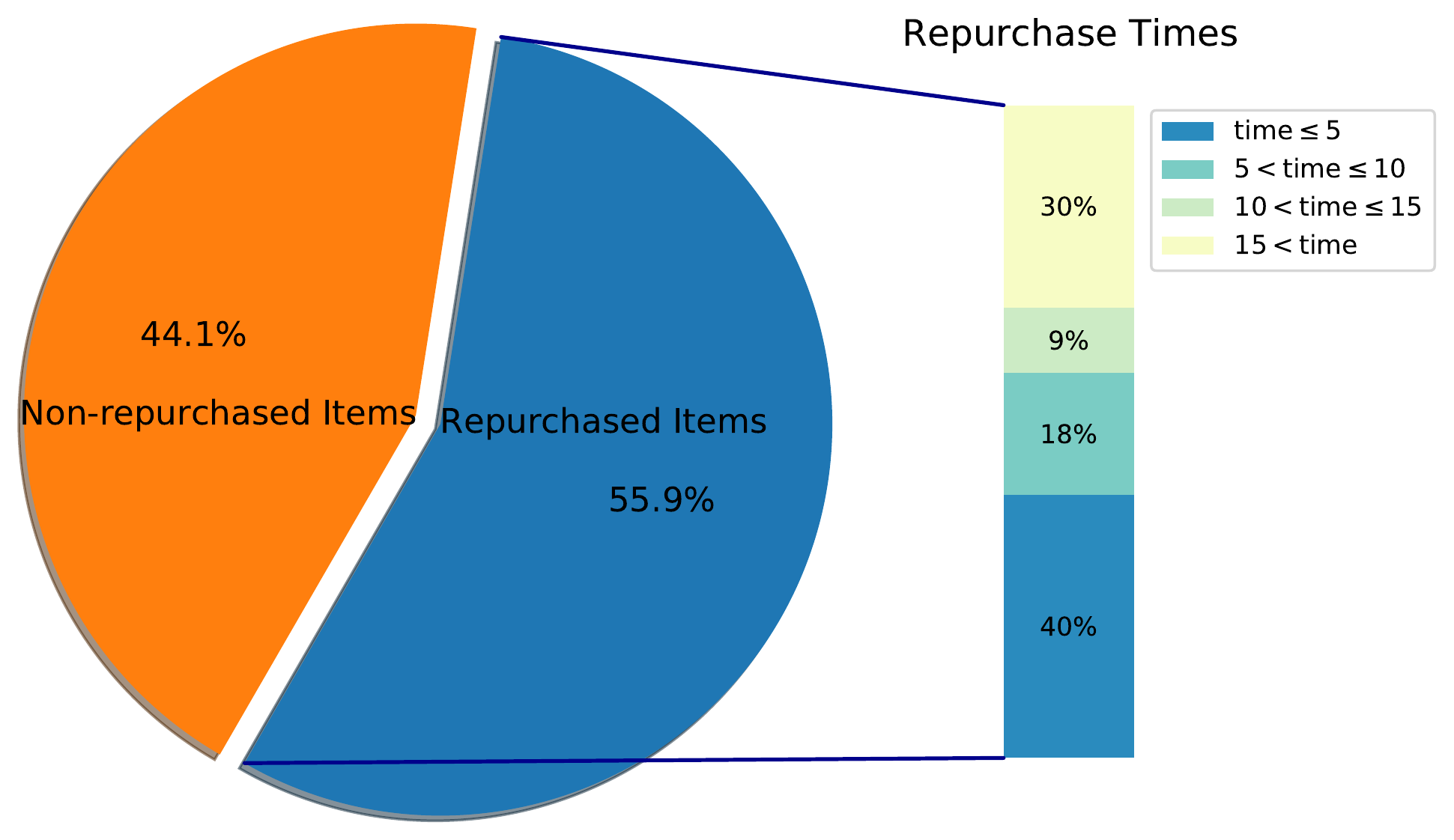}}
    \caption{Item Purchase Behavior Description}
    \label{fig:fig.4}
  \end{minipage}
\end{figure}

The data set used in this section is the user purchase records of T-APP software (hereinafter referred to as T-APP) from August 2019 to September 21, 2020, totalling 89,525 items; In this data set, there are 3321 users and 3245 kinds of commodities. The description of some commodities is shown in Tab.\ref{tab:table1}. The characteristics of each piece of data include user ID, commodity ID, purchase time, purchase quantity, payment method and other attributes. Label the < user, product > repurchase behaviour within the previous time period with the consumption record of one month after August 21, 2020 (forecast period) (i.e., the label with the purchase behaviour occurring before the forecast period and the repurchase behaviour occurring during the forecast period is a positive example; otherwise, it is a negative example.). A total of 2156 users purchased the product during the training period, and 64.92\% of the users had the characteristics of repeated purchase, as shown in Fig.\ref{fig:fig.3}. As can be seen from Fig.\ref{fig:fig.3}, 57\% of users who repurchase have repurchased more than five times. Among all commodity items (a total of 3245 items) purchased by T-App users, 1762 commodity ids are repurchased by users, accounting for 55.86\%, as shown in Fig.\ref{fig:fig.4}.
\begin{table}
	\caption{T-APP product information (partly)}
	\centering
	\begin{tabular}{ll}
		\toprule
		\cmidrule(r){1-2}
		\textbf{ID}     & \textbf{Contents}  \\
		\midrule
		2      & Golden Fried Fish    \\
		38     & Magic sugar cookies   \\
		61     & Magic sugar cookies2 \\
		68     & Pork and celery pie   \\
		69     & Pork and cabbage pie   \\
		73     & Pitaya pie             \\
		\bottomrule
	\end{tabular}
	\label{tab:table1}
\end{table}

In this paper, the user consumption records were extracted by feature extraction, and the training set and verification set were divided by 8:2. The model was trained on the training set, and the training results were fed back on the verification set. Due to the small number of positive samples, random sampling was selected to construct the test set. Model tests were conducted on the test set. The average of the detection results obtained after ten times sampling was taken as the final test result. In the original training, sample set obtained, there are 55,446 negative cases (that is, users do not repurchase the goods) and 529 positive cases (users do repurchase the goods), accounting for 9\% of the positive cases, which are seriously unbalanced samples. In the sample set, use the SMOTE-ENN method for data balance, get 52036 negative samples, 52512 positive samples, the proportion of about 1:1 as the input training sample.

\subsection{Evaluation Methods}
In this paper, accuracy $P$, recall $R$ and $F_1$ score are used to evaluate the performance of the model. In each category of samples, according to the true category and the predicted category of samples, the samples are divided into four types: true cases ($TP$), false-positive cases ($FP$), true negative cases ($TN$), and false-negative cases ($FN$). The confusion matrix is shown in Tab.\ref{tab:table2}. The larger the predicted value is on the main diagonal, and the smaller the value is on the auxiliary diagonal, the better the model. After the confusion matrix is digitized, it is the value of accuracy $P$, recall rate $R$ and $F_1$, and the calculation formula is as follows:
\begin{equation}
    P=\frac{TP}{TP+FP}
\end{equation}
\begin{equation}
    R=\frac{TP}{TP+FN}
\end{equation}
\begin{equation}
    F_1=\frac{2\times{P}\times{R}}{P+R}
\end{equation}

\begin{table}[]
\caption{Prediction results confusion matrix}
\centering
\begin{tabular}{|c|c|c|}
\hline
                                                                                                        & \multicolumn{2}{c|}{\textit{\textbf{Predicted Results}}}                                                                                                          \\ \hline
\textit{\textbf{Real Situation}}                                                                        & \begin{tabular}[c]{@{}c@{}}Positive Example \\ (Repurchase User)\end{tabular} & \begin{tabular}[c]{@{}c@{}}Negative example \\ (Non-repurchase User)\end{tabular} \\ \hline
\begin{tabular}[c]{@{}c@{}}Positive Example \\ (Repurchase User)\end{tabular}                           & $TP$                                                                            & $FN$                                                                                \\ \hline
\multicolumn{1}{|l|}{\begin{tabular}[c]{@{}l@{}}Negative Example\\  (Non-repurchase user)\end{tabular}} & $FP$                                                                            & $TN$                                                                                \\ \hline
\end{tabular}
\label{tab:table2}
\end{table}

\section{Results and Discussion}
\subsection{Model Optimization}
In order to explore the role of the TPE method in model parameter optimization, first use grid search, stochastic search and TPE method respectively to test on the training set after SMOTE-ENN method for sample balance. The search range of parameters of the specified stochastic forest model is shown in Tab.\ref{tab:table3}.
\begin{table}
	\caption{Random forests model parameter space}
	\centering
	\begin{tabular}{lll}
		\toprule
		\cmidrule(r){1-2}
		\textbf{Name}     & \textbf{Range} & \textbf{Counts} \\
		\midrule
		Forest size      & (0,500)    & 10  \\
		Maximum depth of decision tree     & (10,30)  & 20   \\
		Evaluation criteria    & "gini", "entropy"  & 2   \\
		Maximum feature number    & (1,5)  & 4   \\
		\bottomrule
	\end{tabular}
	\label{tab:table3}
\end{table}

\begin{figure}
    \centering
    \fbox{\includegraphics[width = 0.9\textwidth]{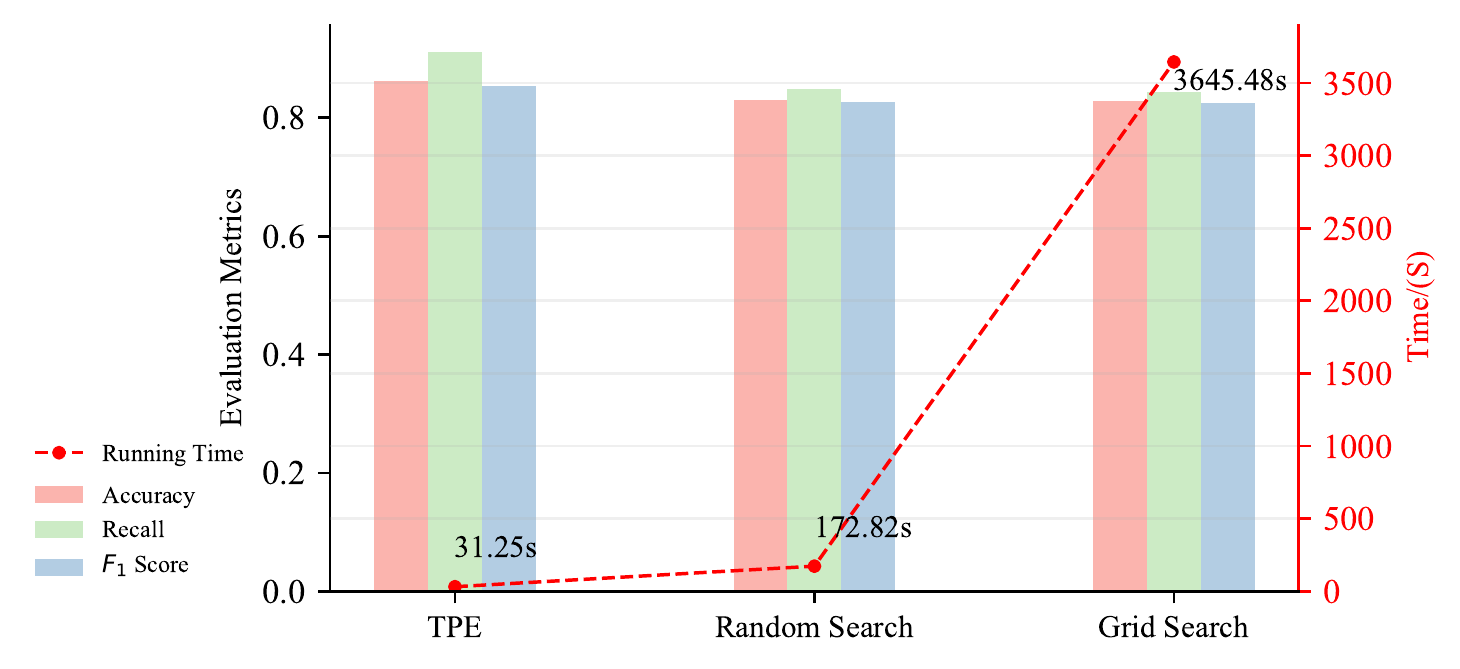}}
    \caption{Comparison of results of different hyperparameter optimization algorithms}
    \label{fig:fig.5}
\end{figure}

Random search and grid search invoke standard machine learning library SciKit-learn 0.24.1\citep{pedregosa2011scikit}. The maximum iteration Times of TPE and random search are specified as 30, and the performance of the model constructed by each method under optimal parameters is shown in Fig.\ref{fig:fig.5}. Times represents the running time of the method.

In Fig.\ref{fig:fig.5}, the random forest model trained by TPE method only takes 31.25s, which is much lower than the random search method and grid search method. Compared with these two methods, the training speed is improved by about 450\% at least, and the accuracy, recall rate and F1 evaluation of the model are higher than those of random search method and grid search method. The reason is that TPE method can form prior knowledge between model parameters and model prediction performance, which can be used to find the optimal parameter structure in each iteration. The above results prove that the TPE method adopted in this paper has the advantages of low time cost and more reasonable model parameter combination.

\subsection{Comparison of Single Models}
The TPE method is used to optimize the hyperparameters of each model, and the values obtained and parameter definitions are shown in Tab.\ref{tab:table4}.
\begin{table}[]
\caption{The main parameters and values of each model}
\centering
\begin{tabular}{cll}
\hline
\textit{\textbf{Model}}                    & \multicolumn{1}{c}{\textit{\textbf{hyper-parameter}}} & \multicolumn{1}{c}{\textit{\textbf{Values}}} \\ \hline
\multirow{5}{*}{\textit{XGBoost}} & Learing\_rate                                         & 0.2990                                       \\
                                           & Depth of trees                                        & 10                                           \\
                                           & Number of trees                                       & 181                                          \\
                                           & Percentage of random sample                           & 0.9740                                       \\
                                           & Penalty term for complexity                           & 3                                            \\ \hline
\multirow{4}{*}{Random Forests}            & Number of trees                                       & 200                                          \\
                                           & Depth of trees                                        & 11                                           \\
                                           & Evaluation criteria                                   & ‘Gini’                                       \\
                                           & Maximum feature number                                & 3                                            \\ \hline
\multirow{4}{*}{LightGBM}                  & Incremental learning                                  & ‘gbdt’                                       \\
                                           & Number of trees                                       & 197                                          \\
                                           & Learing\_rate                                         & 0.25                                         \\
                                           & Maximum number of leaves                              & 31                                           \\ \hline
\end{tabular}
\label{tab:table4}
\end{table}

\begin{figure}
    \centering
    \fbox{\includegraphics[width = 0.75\textwidth]{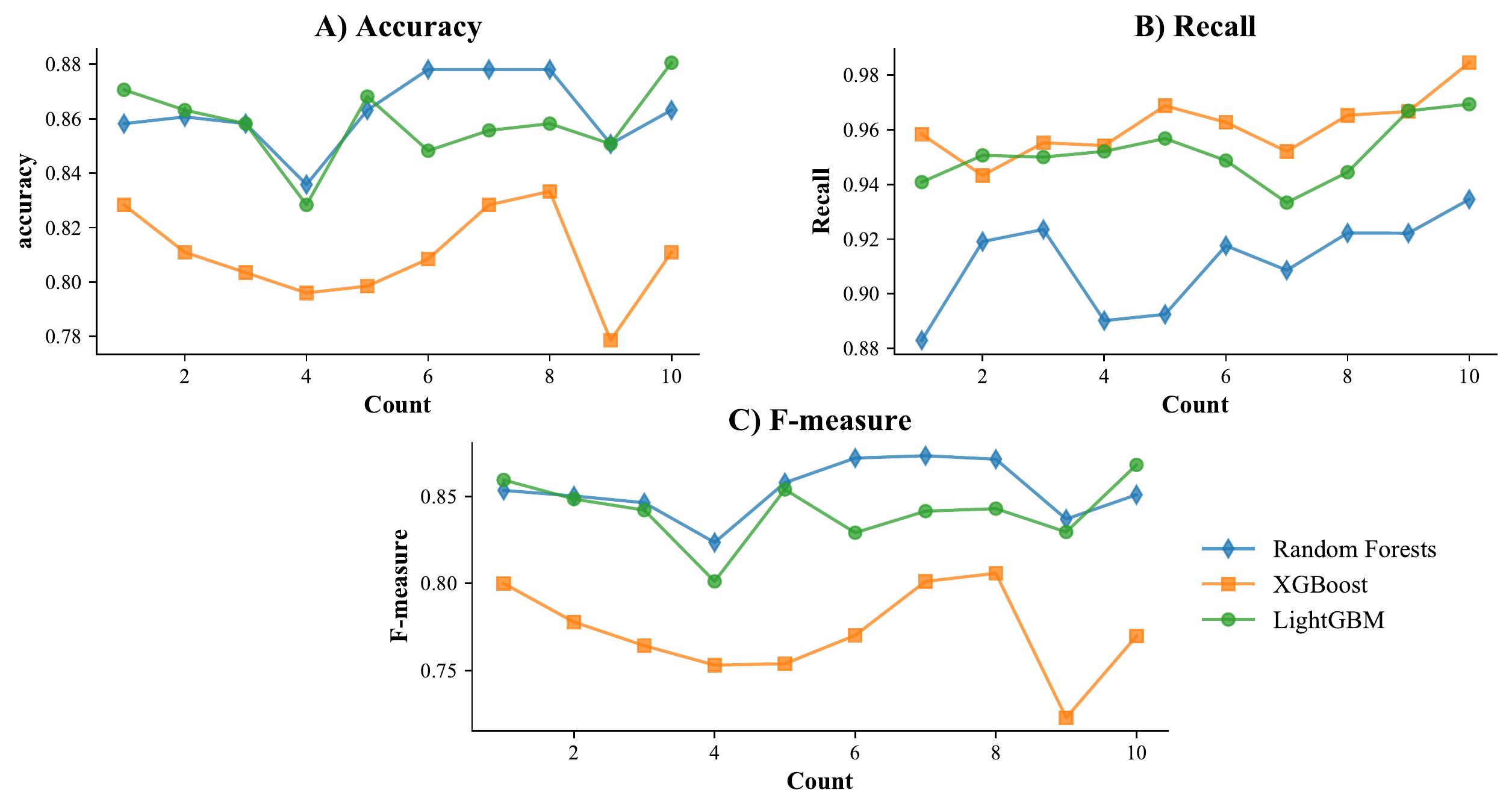}}
    \caption{Single model prediction performance comparison}
    \label{fig:fig.6}
\end{figure}

Ten times random replacement sampling method was used to sample the original positive and negative samples respectively to form the test set, and the prediction performance of each model was obtained as shown in Fig.\ref{fig:fig.6}.

\begin{figure}
    \centering
    \fbox{\includegraphics[width = 0.75\textwidth]{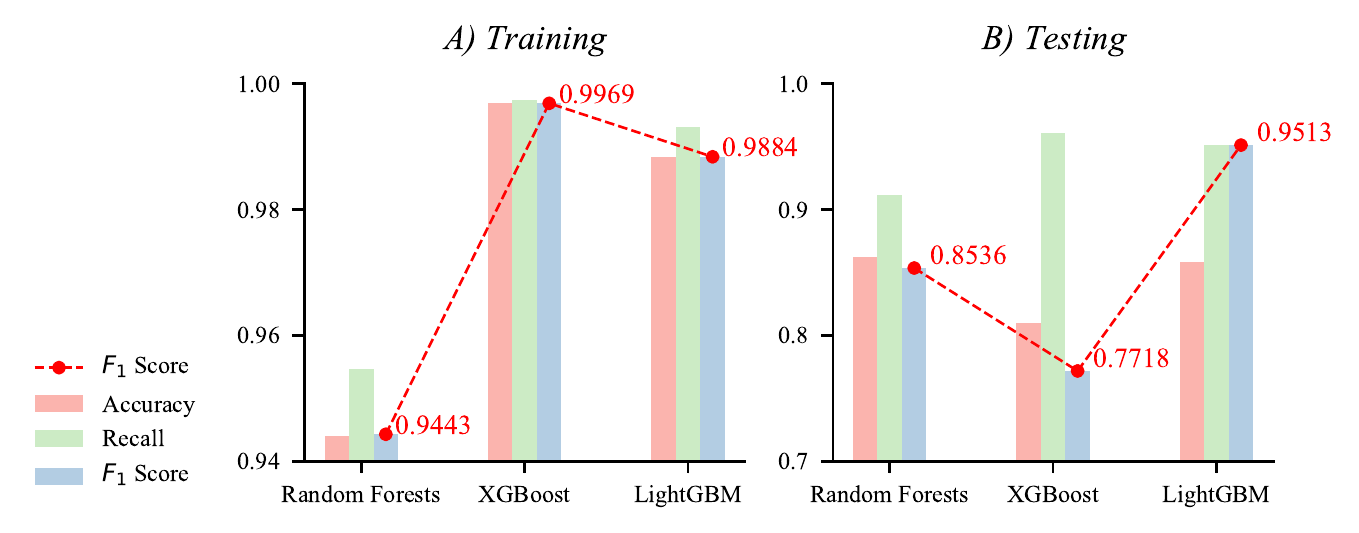}}
    \caption{The average of ten prediction results of each single model}
    \label{fig:fig.7}
\end{figure}

\begin{figure}
    \centering
    \fbox{\includegraphics[width = 0.75\textwidth]{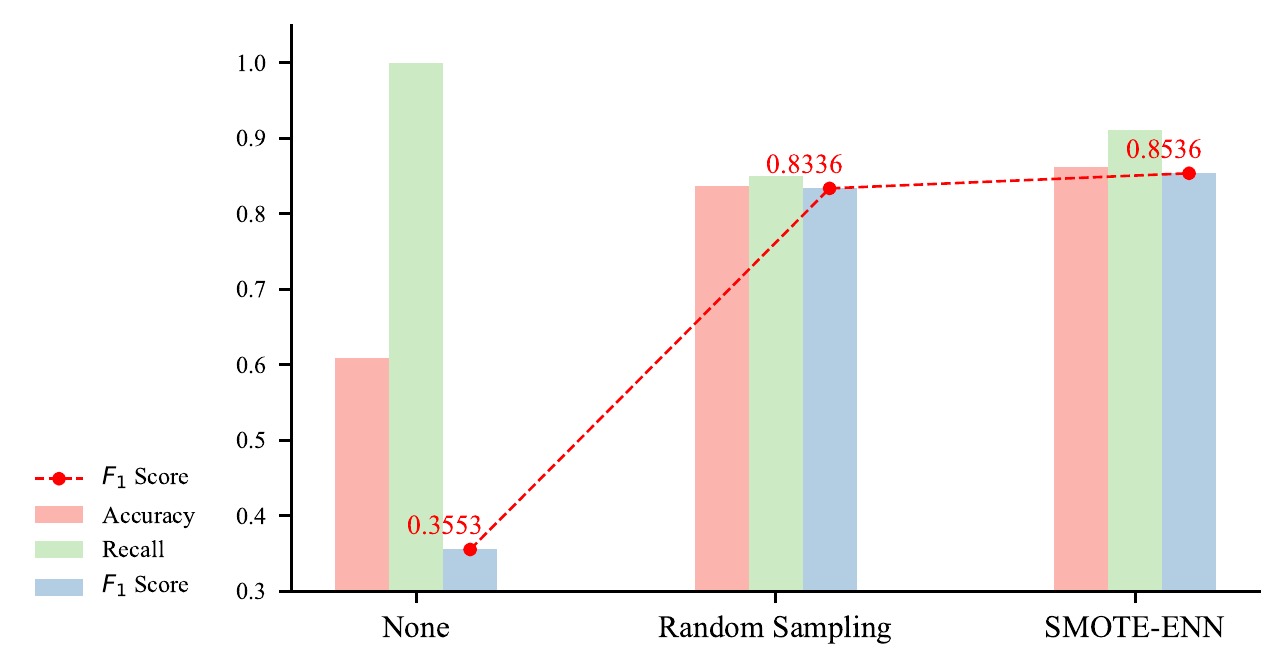}}
    \caption{Comparison of sample balance method results}
    \label{fig:fig.8}
\end{figure}

Figure \ref{fig:fig.7} is obtained by averaging the test results. It can be seen that among the three single models, the random forest model has the strongest prediction performance, with the accuracy, recall rate and $F_1$ score both greater than 0.85 in the test set, followed by the LightGBM model, but the difference between the prediction performance and the random forest model is small. XGBoost model has the lowest predictive performance. The sample balance method (500 negative cases randomly selected, positive and negative ratio 1:1) and the sample balance method are used, respectively. The random forest algorithm is used for ten experiments, and the average value is taken. The prediction performance of each model obtained by comparing with this method on the test set is shown in Figure \ref{fig:fig.8}.

In Figure \ref{fig:fig.8}, we can see that the SMOTE-ENN method used in this paper can improve the prediction performance of the model.

\subsection{Model Integration}

The soft voting method mentioned above is used to integrate the random forest algorithm and LightGBM algorithm with better performance in this paper (both models have the best hyperparameters). Bagging ensemble method is used for random forest algorithm and Boosting ensemble method is used for LightGBM. Further integration of random forest algorithm and LightGBM with soft voting method can make full use of the advantages of the two ensemble methods and achieve better prediction performance.

The soft Voting method uses the Voting Classifier module provided by Scikit-Learn, a Python-based open source machine learning platform. Under the same training set, RF-LightGBM model is trained and tested, and the results obtained are shown in Table \ref{tab:table5}.

\begin{table}[]
\caption{Comparison between Integration model and single model}
\centering
\begin{tabular}{|c|c|c|c|c|c|c|c|}
\hline
\multicolumn{1}{|l|}{} & \multicolumn{3}{c|}{Train set} & \multicolumn{3}{c|}{Test Set} & Reference                  \\ \hline
\multicolumn{1}{|l|}{} & Accuracy   & Recall   & F1     & Accuracy   & Recall  & F1     &                             \\ \hline
\textcolor{red}{RF}                     & 0.944      & 0.955    & 0.944  & 0.862      & 0.911   & 0.853  & \multirow{3}{*}{\textcolor{red}{This Paper}} \\ \cline{1-7}
\textcolor{red}{LightGBM}               & 0.988      & 0.993    & 0.988  & 0.858      & 0.951   & 0.842  &                             \\ \cline{1-7}
\textcolor{red}{RF-LightGBM}            & 0.987      & 0.991    & 0.987  & 0.871      & 0.952   & 0.859  &                             \\ \hline
LSTM                   & 0.828      & 0.854    & 0.841  & 0.802      & 0.842   & 0.822  & \multirow{2}{*}{\citep{huxiaoli2020baed}}   \\ \cline{1-7}
CNN-LSTM               & 0.875      & 0.843    & 0.856  & 0.856      & 0.839   & 0.847  &                             \\ \hline
Teacher-Student        & 0.9193     & Non      & Non    & 0.9198     & Non     & Non    & {\citep{shen2020reconciling}}                     \\ \hline
\end{tabular}
\label{tab:table5}
\end{table}

It can be seen that the RF-LighTGBM integrated model constructed in this paper has a better prediction effect than a single model. Although the increase of F1 value is not high, it is not an increase that can be ignored considering that the prediction problem is more demanding for F1 value. The performance in the training set of the model in this paper is greater than that in works of \citep{shen2020reconciling} and \citep{huxiaoli2020baed}, indicating that the SMOTE-ENN method used in this paper can train the model better than its method. In addition, the RF-lightGBM integrated model constructed in this paper has a better F1 score and higher prediction accuracy than the model constructed in reference \citep{shen2020reconciling} and \citep{huxiaoli2020baed}.

\section{Conclusion}
Compared with full-category e-commerce platforms, community e-commerce has the characteristics of relatively fixed user groups, relatively single consumer goods, and apparent characteristics of repeated purchase. Research on the prediction of repurchase behaviour of community e-commerce users can effectively guide the operational decisions of community e-commerce platform operators. This paper conducts a more in-depth study on the previous data-driven methods. The results show that: (1) the improved RFM model proposed in the feature engineering of this paper can effectively represent the repurchase behaviour of community e-commerce customers; (2) The method proposed in this paper using SMOTE-ENN for predicting the training sample balance of customer repurchase behaviour can more effectively complete the training of machine learning model; (3) Compared with other methods, an automatic hyperparameter optimization method based on TPE method proposed in this paper can significantly improve the model training efficiency and model prediction performance, and the training time is reduced by more than 450\%; (4) The RF-LighTGBM model proposed in this paper can effectively predict customer repurchase behaviour and is superior to the single model and previous research results.

The possible deficiency of this work is the lack of user data. Limited by the management information system of the community e-commerce platform, this paper did not obtain rich user information (such as gender, age, etc.), and these demographic characteristics may have a significant impact on consumers' purchasing behaviour so as to train a more accurate prediction model. Future research can focus on obtaining more detailed input data, looking for features that can more accurately depict customer buying behaviour, and using better methods to solve problems such as sample imbalance and complex parameter optimization in the prediction process.

\bibliographystyle{unsrtnat}
\bibliography{references}  






\end{document}